\documentclass[conference]{IEEEtran}
\IEEEoverridecommandlockouts
\usepackage{cite}
\usepackage{amsmath,amssymb,amsfonts}
\usepackage{placeins}
\usepackage{algorithm}
\usepackage{algpseudocode}
\usepackage{graphicx}
\usepackage{subcaption}
\usepackage{textcomp}
\usepackage{xcolor}
\usepackage{multirow}

\def\BibTeX{{\rm B\kern-.05em{\sc i\kern-.025em b}\kern-.08em    T\kern-.1667em\lower.7ex\hbox{E}\kern-.125emX}}

\usepackage{tikz}
\usetikzlibrary{graphs,arrows,positioning,trees,decorations.pathmorphing,chains,shapes,decorations.pathreplacing,quotes}
\definecolor{myBlue}{RGB}{0,117,189}
\definecolor{myRed}{RGB}{217,83,25}
\definecolor{myYellow}{RGB}{237,177,32}

\tikzstyle{hidden} = [circle,draw]
\tikzstyle{visible} = [circle,draw,fill=black!10]
\tikzstyle{myLink} = [->,>=triangle 45]
\tikzstyle{MyArrow} = [single arrow, fill=red, opacity=0.2, text opacity=1, minimum width=1cm, minimum height=3cm, inner sep=0.1cm, single arrow head extend=1ex]

\graphicspath{{plots/}}

\begin{document}
    
    \title{Bayesian Tensor Factorisation for Bottom-up Hidden Tree Markov Models %    Tensor Factorisation Bottom-Up Hidden Tree Markov Model
    }
    \author{\IEEEauthorblockN{Daniele Castellana and Davide Bacciu}
\IEEEauthorblockA{Dipartimento di Informatica\\
Universit\`a di Pisa\\
Largo B. Pontecorvo 3\\
Pisa, Italy\\
Email: \{daniele.castellana,bacciu\}@di.unipi.it}
}

    \maketitle
    
    \begin{abstract}
        Bottom-Up Hidden Tree Markov Model is a highly expressive model for tree-structured data. Unfortunately, it cannot be used in practice due to the intractable size of its state-transition matrix. We propose a new approximation which lies on the Tucker factorisation of tensors. The probabilistic interpretation of such approximation allows us to define a new probabilistic model for tree-structured data. Hence, we define the new approximated model and we derive its learning algorithm. Then, we empirically assess the effective power of the new model evaluating it on two different tasks. In both cases, our model outperforms the other approximated model known in the literature.
    \end{abstract}

    \section{Introduction}
    %Tree-structured data arise in multiple contexts to represent hierarchically-organized information. For example, in Natural Language Processing, a sentence can be represented as a tree where leaf and internal nodes represent respectively words and syntactic category. In the web, most of the information is represented through tree structures thanks to the diffusion of the XML standard. Regardless of the domain application, developing a Machine Learning model which is able to deal with such rich representation pose two main challenges \cite{Frasconi1998}: (1) the model must be able to adapt itself on different tree structures and (2) the model must exploit the structure information to absolve the given task.
    
    %The former challenge is met building a recursive model. For example, the Tree-LSTM \cite{Tai2015} take advantage of the recursive definition of the Long Short-Term Memory (LSTM) unit firstly used on sequences. Also, Hidden Markov Models (HMMs) provide a recursive framework (firstly used on sequence) which can be useful to deal with tree-structured data (e.g. \cite{Bacciu2012,Crouse1998}).
    
    %The latter challenge is more complex. Given a tree-structured data, the structure naturally defines a context for each node: the set of surrounding nodes. The context size has a direct impact on the search space of our algorithm. In fact, the number of possible dependency between a node and its context grows exponentially with respect to the size of the context.
    Trees are complex data which represent hierarchical information. They are composed of atomic entities, called nodes, combined together through a parent-child relationship which defines the tree structure. Developing a Machine Learning model which is able to deal with such rich representation poses two main challenges \cite{Frasconi1998}: (1) the model must be able to adapt itself on different tree structures and (2) the model must exploit the structure information to absolve the given task.
    
    Both challenges have been tackled over the years by a variety of approaches ranging from early works on recursive neural networks \cite{Frasconi1998}, to their more recent deep learning style re-factoring \cite{treelstmbu} (see \cite{inns2019} for a recent survey), and including kernel-based approaches \cite{DBLP:journals/sigkdd/Gartner03,Shin2014,genKerTNNLS} and generative models \cite{Crouse1998,Bacciu2012,gtmsd}. 
    In this work, we focus in particular on the latter models, which are referred to as Hidden Tree Markov Models (HTMM), being obtained as a generalisation of the Hidden Markov Model (HMM) to the tree domain. Like their sequential counterpart, HTMM defines a generative process for tree structures regulated by discrete-valued hidden state variables, typically from a finite alphabet (but can be extended to non-parametric models along the lines of \cite{neucomBayesHTMM}). Structural knowledge is captured by such hidden states through their transition distribution, whose representational power and complexity depend on the direction of context propagation. In the literature, we mainly refer to two types of HTMM associated with different context propagation strategies. The Top-Down Hidden Tree Markov Model (TD-HTMM) \cite{Crouse1998} defines a generative process propagating from the root to the leaves of the tree. The top-down direction defines a causal relationship among nodes which goes from the parent to each child independently. On the other hand, the Bottom-Up Hidden Tree Markov Model (BU-HTMM) \cite{Bacciu2012} defines a generative process which goes in the opposite direction: from the leaves to the root. The different direction induces more complex causal relationships between nodes, pointing from the joint state of the children towards the common parent. Different casual relationships create differences in the local Markov properties of the two approaches, which influence the way the nodes exchange information during inference and learning \cite{Bacciu2012}.
    
    While possessing several interesting properties, e.g. compositionality \cite{gtmsd}, the bottom-up HTMM has practical limitations due to the size of the state-transition distribution which is exponential with respect to the output degree (i.e. the maximum number of child nodes). To overcome this limitation, the authors in \cite{Bacciu2012} introduced an approximation, named Switching Parent (SP), which factorises the complex joint state transition distribution as a mixture of simpler multinomials. While having demonstrated its effectiveness in a number of applications, including generalisation to graphs \cite{icml2018}, such an approximation limits the amount of information shared between the children of common parents.
    
    The goal of this paper is to introduce a new approximation for the BU-HTMM state-transition distribution which relies on a tensor decomposition known as Tucker decomposition \cite{Tucker1966}. This approximation is interesting because of its probabilistic interpretation, highlighted in \cite{Yang2016,Sarkar2018}, as well as for the increased information sharing between children with respect to the SP factorisation.  Such an advantage comes at the cost of a more complex learning algorithm, due to the use of Gibbs sampling when estimating model parameters from data. As part of this work, we validate our proposed Bayesian tensor factorisation approach on two different learning tasks. The former is a standard tree classification task, while the latter tests the model ability to predict node labels given a tree structure. In both tasks, we compare our model to the BU-HTMM model with SP approximation introduced in \cite{Bacciu2012}. Results highlight how the proposed tensor factorisation outperforms the SP model, especially in the label prediction task.
    
    The rest of the paper is organised as follows. In Section \ref{sec:background}, we introduce the two building blocks of our new model. In Section \ref{sec:model} we formally introduce our model and we derive its learning algorithm. In Section \ref{sec:experiment} we report the empirical analysis, while in Section \ref{sec:conclusion} we draw our conclusions.
    
    \section{Background}\label{sec:background}
    \subsection{Bottom-Up Hidden Tree Markov Model}
    The Bottom-Up Hidden Tree Markov Model (BHTMM) \cite{Bacciu2012} defines a probability distribution over tree-structured data by postulating the existence of a hidden generative process regulated by unobserved Markov state random variables.
    
    The terms Bottom-Up refers to the direction of the conditional dependence among tree nodes. In particular, the BHTMM assumes that the hidden state of a node depends on the joint hidden state of its direct descendants in the tree.
    
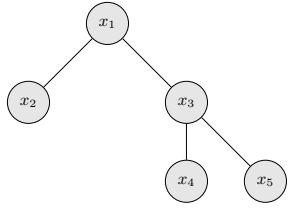
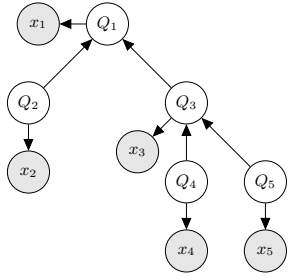
\begin{figure}[tb]
        \centering
        \begin{subfigure}[t]{0.48\textwidth}
            \centering
            \small
            \scalebox{0.7}{
                \begin{tikzpicture}[edge from parent/.style={draw,-}, every node/.style = {minimum size = 0.8cm}, execute at begin node=$, execute at end node=$,node distance = 0.1cm]
                \node [visible] (q1) {x_1}
                child {node [visible] (q2) {x_2}}
                child [missing]
                child {node [visible] (q3) {x_3}
                    child [missing]
                    child {node [visible] (q4) {x_4}}
                    child {node [visible] (q5) {x_5}}
                };
                \end{tikzpicture}%
            }
            \caption{\footnotesize The observed tree $\mathbf{x}$}
        \end{subfigure}
        
        \vspace{10pt}
        
        \begin{subfigure}[t]{0.48\textwidth}
            \centering
            \small
            \scalebox{0.7}{
                \begin{tikzpicture}[edge from parent/.style={draw,<-,>=triangle 45}, every node/.style = {minimum size = 0.8cm}, execute at begin node=$, execute at end node=$,node distance = 0.5cm]
                \node [hidden] (q1) {Q_1}
                child {node [hidden] (q2) {Q_2}}
                child [missing]
                child {node [hidden] (q3) {Q_3}
                    child [missing]
                    child {node [hidden] (q4) {Q_4}}
                    child {node [hidden] (q5) {Q_5}}
                };
                \node [visible] (y1) [left=of q1] {x_1};
                \node [visible] (y2) [below=of q2] {x_2};
                \node [visible] (y3) [below left=of q3] {x_3};
                \node [visible] (y4) [below=of q4] {x_4};
                \node [visible] (y5) [below=of q5] {x_5};
                
                %\node[hidden] (s3) [above right= of q3] {S_3};
                %\node[hidden] (s1) [above right= of q1] {S_1};
                
                \draw [myLink] (q1) to (y1);
                \draw [myLink] (q2) to (y2);
                \draw [myLink] (q3) to (y3);
                \draw [myLink] (q4) to (y4);
                \draw [myLink] (q5) to (y5);
                
                %\draw [myLink] (s1) to (q1);
                %\draw [myLink] (s3) to (q3);
                
                \end{tikzpicture}%
            }
            \caption{\footnotesize The BHTMM graphical model corresponding to the tree above.}
        \end{subfigure}
        \caption{An observed tree $\mathbf{x}$ (a) and its associated BHTMM graphical model (b).}
        \label{GM_BHTMM}
    \end{figure}

    Given a labelled tree $t$, we denote with $\mathbf{x} =\{x_1,\dots, x_{|t|}\}$ the set of visible labels in $t$. Then, its BHTMM graphical model is built associating a discrete hidden random variable $Q_u \in [1,C]$ to each label $x_u$ in the tree. All hidden variables are linked together reproducing the same structure of the visible tree $t$; the direction of links goes from leaves to the root in order to represent the dependence between the hidden state of a node and the joint configuration of its hidden child nodes (see Fig. \ref{GM_BHTMM}).
    
    Given the graphical model, the derivation of the complete likelihood associated to labels $\mathbf{x}$ is straightforward:
    \begin{equation}
    \begin{split}
    &\mathcal{L}(\mathbf{x}, \mathbf{Q})  = P(\mathbf{x}, \mathbf{Q}) =
    \prod_{u \in \mathcal{LF}} \pi^l(j) \,b_j(x_u)\\
    &\quad \times \prod_{u \notin \mathcal{LF}} A_{j_1,\dots,j_L}(j)\, b_j(x_u),    
    \end{split}
    \label{eq:BTHMM_like}
    \end{equation}
    where we use $\mathbf{Q}=\{Q_1,\dots,Q_{|t|}\}$ to indicate the set of the hidden variables introduced by the BHTMM. The term $L$ is the maximum tree output degree (i.e. the maximum number of children).
    
    The distributions in \eqref{eq:BTHMM_like}, as in every HMM, are: the \textit{priori} distribution $\pi^l(j) = P^l(Q_u=j)$, the \textit{state-transition} distribution $A_{j_1,\dots,j_L}(j) = P(Q_u =j \mid Q_{ch_1(u)} = j_1, \dots, Q_{ch_L(u)}=j_L)$, and the \textit{emission} distribution $b_j(x_u) = P(x_u \mid Q_u=j)$. In our case, the priori distribution is defined on leaf nodes: we use the symbol $\mathcal{LF}(t)$ to indicate the set of leaf nodes in $t$. Moreover, we use a different priori distribution for each node position: the $l$ superscript indicates the positional dependence of the priori distribution. To our purpose, the position of a node is computed with respect to its siblings: we denote by $ch_l(u)$ the $l$-th child node of $u$ and by $l=pos(u)$ the position of node $u$ w.r.t. its siblings. We will omit the dependence on $u$ when it clear from the context or it is not relevant.
    
    The state-transition distribution models the relationships between the hidden state of a node and the joint hidden state of its child nodes; unfortunately, this relationship cannot be represented explicitly. In fact, given $C$ the number of hidden states and $L$ the maximum tree output degree, the state-transition distribution is represented by a tensor which contains $C^{L+1}$ elements. The exponential growth with respect to the number of child nodes $L$ is a direct consequence of the assumption made by the model since there is an exponential number of joint state configurations.
    
    To this end, an approximation is required to make the Bottom-Up approach feasible in practice. The first approximation introduced is called \textit{Switching-Parent} \cite{Bacciu2012}: the idea is to approximate of the joint transition distribution as a mixture of simpler distributions. In practice, such approximation assumes the existence of a mixture variable $S_u$ for each tree internal nodes. Then, the joint transition matrix factorises as a convex combination of $L$ elementary transition matrices (one for each child), where the mixture distribution $P(S_u=l)$ represents the weight of such combination. In formula, the following simplification is applied:
    \begin{equation}
    \begin{split}
    &P(Q_u =j \mid Q_{ch_1(u)}=j_1, \dots, Q_{ch_L(u)}=j_L)\\
    &\qquad =\sum_{l=1}^L P(S_u=l) \, P(Q_u =j\mid S_u = l, Q_{ch_l(u)}=j_l).
    \end{split}
    \end{equation}
    This approximated model is referred to as SP-BHTMM.
    
    \subsection{Tucker Decomposition} \label{sec:tucker_decomp}
    Let $\mathcal{A} \in \mathbb{R}^{I_1 \times \dots \times I_p}$ a $p$-way tensor, each element $a_{x_1,\dots, x_p}$ in $\mathcal{A}$ can be factorised as \cite{DeLathauwer2000}
    \begin{equation}
    a_{x_1,\dots, x_p} = \sum_{h_1=1}^{J_1} \cdots \sum_{h_p=1}^{J_p} \lambda_{h_1,\dots,h_p} \prod_{z=1}^p \kappa^{(z)}_{h_z,x_z},
    \label{eq:Tucker_decomp}
    \end{equation}
    where $\lambda_{h_1,\dots,h_p}$ are elements of a new $p$-way tensor $\boldsymbol{\lambda} \in \mathbb{R}^{J_1 \times \dots \times J_p}$ called \textit{core} tensor and $\kappa^{(z)}_{h_z,x_z}$ are elements of matrices $\boldsymbol{\kappa}^{(z)} \in \mathbb{R}^{I_z \times J_z}$, called \textit{mode} matrices.
    
\begin{figure}[tb]
        \centering
        \begin{subfigure}[t]{0.48\textwidth}
            \centering
            \begin{tikzpicture}[every edge quotes/.append style={auto, text=blue}]
            \pgfmathsetmacro{\cubex}{2.4}
            \pgfmathsetmacro{\cubey}{2}
            \pgfmathsetmacro{\cubez}{1.6}
            \pgfmathsetmacro{\corex}{1.2}
            \pgfmathsetmacro{\corey}{1.1}
            \pgfmathsetmacro{\corez}{1}
            \draw [draw=black, every edge/.append style={draw=black, densely dashed, opacity=.5}, fill=blue!30]
            (0,0,0) coordinate (o) -- ++(-\cubex,0,0) coordinate (a) -- ++(0,-\cubey,0) coordinate (b) edge coordinate [pos=1] (g) ++(0,0,-\cubez)  -- ++(\cubex,0,0) coordinate (c) -- cycle
            (o) -- ++(0,0,-\cubez) coordinate (d) -- ++(0,-\cubey,0) coordinate (e) edge (g) -- (c) -- cycle
            (o) -- (a) -- ++(0,0,-\cubez) coordinate (f) edge (g) -- (d) -- cycle;
            \path [every edge/.append style={draw=blue, |-|}]
            (b) +(0,-5pt) coordinate (b1) edge ["$I_2$"'] (b1 -| c)
            (b) +(-5pt,0) coordinate (b2) edge ["$I_1$"] (b2 |- a)
            (c) +(3.5pt,-3.5pt) coordinate (c2) edge ["$I_3$"'] ([xshift=3.5pt,yshift=-3.5pt]e);
            \node at (-\cubex/2,-\cubey/2,-\cubez/2) {$\mathcal{A}$};
            \end{tikzpicture}
            \caption{The tensor $\mathcal{A}$.}
        \end{subfigure}
        
        \begin{subfigure}[t]{0.48\textwidth}
            \centering      
            \begin{tikzpicture}[every edge quotes/.append style={auto, text=blue}]
            \pgfmathsetmacro{\cubex}{2.4}
            \pgfmathsetmacro{\cubey}{2}
            \pgfmathsetmacro{\cubez}{1.6}
            \pgfmathsetmacro{\corex}{1.2}
            \pgfmathsetmacro{\corey}{1.1}
            \pgfmathsetmacro{\corez}{1}
            \draw [draw=black, every edge/.append style={draw=black, densely dashed, opacity=.5}, fill=brown!40]
            (0,0,0) coordinate (o) -- ++(-\corex,0,0) coordinate (a) -- ++(0,-\corey,0) coordinate (b) edge coordinate [pos=1] (g) ++(0,0,-\corez)  -- ++(\corex,0,0) coordinate (c) -- cycle
            (o) -- ++(0,0,-\corez) coordinate (d) -- ++(0,-\corey,0) coordinate (e) edge (g) -- (c) -- cycle
            (o) -- (a) -- ++(0,0,-\corez) coordinate (f) edge (g) -- (d) -- cycle;
            %\path [every edge/.append style={draw=blue, |-|}]
            %\(b) +(0,-5pt) coordinate (b1) edge ["$J_2$"'] (b1 -| c)
            %(b) +(-5pt,0) coordinate (b2) edge ["$J_1$"] (b2 |- a)
            %(c) +(3.5pt,-3.5pt) coordinate (c2) edge ["$J_3$"'] ([xshift=3.5pt,yshift=-3.5pt]e);
            \node at (-\corex/2,-\corey/2,-\corez/2) {$\boldsymbol{\lambda}$};
            
            \draw [draw=black, every edge/.append style={draw=black, densely dashed, opacity=.5}, fill=red,fill opacity=0.3]
            (0,0.5,0) --++ (-\corex,0,0) coordinate (r1) --++ (0,\cubex,0) coordinate (r2) --++(\corex,0,0)  coordinate (r3)-- cycle;
            \path [every edge/.append style={draw=blue, |-|}]
            (r1) +(-5pt,0) coordinate (rrr) edge ["$I_1$"] (rrr |- r2)
            (r2) +(0,5pt) coordinate (rr) edge ["$J_1$"] (rr -| r3);
            \node at (-\corex/2,0.5+\cubex/2,0) {$\boldsymbol{\kappa}^{(1)}$};
            
            \draw [draw=black, every edge/.append style={draw=black, densely dashed, opacity=.5}, fill=green, fill opacity=0.3]
            (0.3,-\corey,0) --++ (0,0,-\corez) coordinate (g1) --++(\cubez,0,0) coordinate (g2) --++ (0,0,+\corez) coordinate (g3) -- cycle;
            \path [every edge/.append style={draw=blue, |-|}]
            (g1) +(0,5pt) coordinate (gg) edge ["$I_3$"] (gg -| g2)
            (g2) +(3.5pt,-3.5pt) coordinate (ggg) edge ["$J_3$"] ([xshift=3.5pt,yshift=-3.5pt] g3);
            \node at (0.3+\cubez/2, -\corey, -\corez/2) {$\boldsymbol{\kappa}^{(3)}$};
            
            \draw [draw=black, every edge/.append style={draw=black, densely dashed, opacity=0.5}, fill=yellow, fill opacity=0.3]
            (-\corex - 0.2,0) --++(0,-\corey,0) coordinate (y1) --++ (-\cubey,0,0) coordinate (y2) --++(0,\corey,0) coordinate (y3) -- cycle;
            \path [every edge/.append style={draw=blue, |-|}]
            (y1) +(0,-5pt) coordinate (yy) edge ["$I_2$"] (yy -| y2)
            (y2) +(-5pt,0) coordinate (yyy) edge ["$J_2$"] (yyy |- y3);
            \node at (-\corex -0.2 - \cubey/2, -\corey/2,0) {$\boldsymbol{\kappa}^{(2)}$};
            \end{tikzpicture}
            \caption{The Tucker decomposition.}
        \end{subfigure}
        \caption{A three-way tensor $\mathcal{A}$ (a) and its Tucker decomposition (b).}
        \label{fig:Tucker_decomp}
    \end{figure}
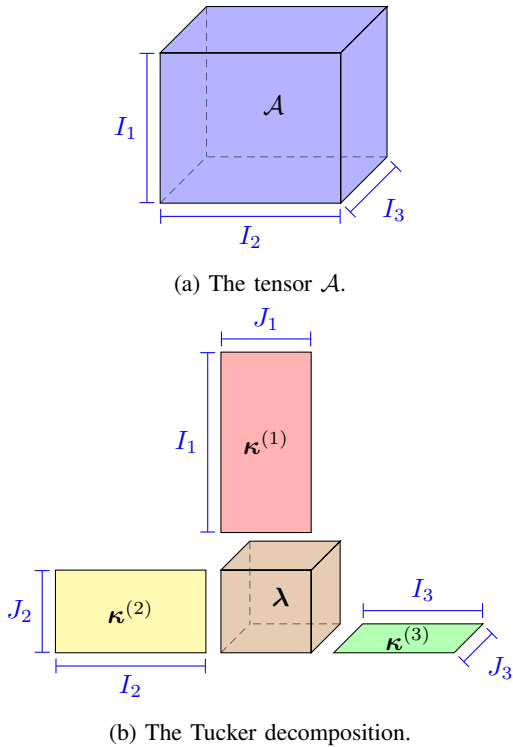

    A more compact way to represent the same decomposition can be obtained using the operator $\times_n$, which represents $n$-mode product of a tensor by a matrix along the $n$-th dimension \cite{DeLathauwer2000}:
    \begin{equation}
    \mathcal{A} = \boldsymbol{\lambda} \times_1 \boldsymbol{\kappa}^{(1)} \times_2 \boldsymbol{\kappa}^{(2)} \times_3 \dots \times_p \boldsymbol{\kappa}^{(p)}.
    \end{equation}
    An example of decomposition can be visualised in Figure \ref{fig:Tucker_decomp}.
    
    This factorisation is called \textit{Tucker} decomposition, since it was originally developed by Tucker to obtain a method for searching relations in a three-way tensor of psychometric data \cite{Tucker1966}. Later, the same idea has been generalised to $p$-way tensor proving that it is a multi-linear generalisation of the matrix Singular Value Decomposition (SVD) \cite{DeLathauwer2000}. Hence, the same decomposition is also called High-Order Singular Value Decomposition (HOSVD) \cite{DeLathauwer2000}.
    
    The major advantage of the Tucker decomposition is that it may lead to a compressed representation of the original tensor. In fact, storing the tensor $\mathcal{A}$ in its explicit form requires $\prod_{z=1}^p I_z$ space. Instead, storing its factorised form requires $\prod_{z=1}^p J_z$ space to store $\boldsymbol{\lambda}$ tensor plus $\sum_{z=1}^p I_z \times J_z$ to store all the $\boldsymbol{\kappa}^{(.)}$ matrices. Thus, the Tucker factorisation is advantageous if and only if $\prod_{z=1}^p J_z \ll \prod_{z=1}^p I_z$.
    
    Even if we have the guarantee that a Tucker decomposition exists for every tensor \cite{DeLathauwer2000}, we have no guarantee on the size of its core tensor. In the worst case, the core tensor has the same dimension as the initial one; hence, there is no compression using the Tucker decomposition.
    
    \section{The Tensor Factorisation BHTMM}\label{sec:model}
    
    \subsection{Model}\label{sec:model_intro}
    The Tensor Factorisation Bottom-up Hidden Tree Markov Model (TF-BHTMM) defines a probability distribution over tree-structured data. As for BHTMM, the direction of conditional dependencies among tree nodes goes from the leaves to the root. Moreover, the Tucker decomposition is applied in order to reduce the complexity required to store a BHTMM state-transition distribution.
    
\begin{figure}[tb]
        \centering
        \scalebox{0.7}{
            \begin{tikzpicture}[edge from parent/.style={draw,<-,>=triangle 45}, every node/.style = {minimum size = 1.1cm}, execute at begin node=$, execute at end node=$,node distance = 0.5cm]
            \node [hidden] (qu) {Q_u}
            child {node [hidden] (z1) {z_{ch_1}}
                child{node [hidden] (q1) {Q_{ch_1}}}
            }
            child {node [draw=none,minimum size = 0.1cm] {\dots}  edge from parent[draw=none]}
            child {node [hidden] (zl) {z_{ch_l}}
                child{node [hidden] (ql) {Q_{ch_l}}}
            }
            child {node [draw=none,minimum size = 0.1cm] {\dots}  edge from parent[draw=none]}
            child {node [hidden] (zL) {z_{ch_L}}
                child{node [hidden] (qL) {Q_{ch_L}}}
            }
            ;
            \node [visible] (yu) [above=of qu] {x_u};
            \node [visible] (y1) [below=of q1] {x_{ch_1}};
            \node [visible] (yl) [below=of ql] {x_{ch_l}};
            \node [visible] (yL) [below=of qL] {x_{ch_L}};
            
            \draw [myLink] (qu) to (yu);
            \draw [myLink] (q1) to (y1);
            \draw [myLink] (ql) to (yl);
            \draw [myLink] (qL) to (yL);
            
            \end{tikzpicture}%
        }
        \caption{Markov blanket of the hidden variable $Q_u$. in the TF-BHTMM approximation.}
        \label{fig:markov_blanket_u}
    \end{figure}
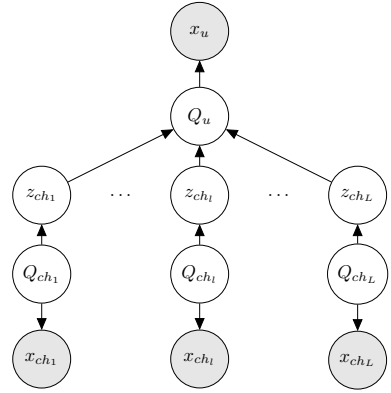

    To this end, given a labelled tree $t$, the TF-BHTMM introduces a new set of discrete hidden random variable $\mathbf{z} = \{z_1,\dots,z_{|t|}\}$. The goal of these variables is to break the dependence between a node and the joint configuration of its children. Hence, the state of a variable $z_u$ depends only on the state of the variable $Q_u$, while the state of a variable $Q_u$ depends on the joint configuration $(z_{ch_1},\dots,z_{ch_L})$ associated to its children. In Figure \ref{fig:markov_blanket_u} we show the Markov blanket associated to a generic tree node $u$.
    By observing this figure, it is clear that the every path between the variable $Q_u$ and every child variable $Q_{ch_l}$ is blocked by the variable $z_{ch_l}$. Hence, we can argue that the variable $Q_u$ is conditionally independent of the joint configuration $(Q_{ch_1}, \dots, Q_{ch_L})$ when the joint configuration $(z_{ch_1},\dots,z_{ch_L})$ is observed.    
    
    The interaction between the $\mathbf{z}$ and the $\mathbf{Q}$ variables is modelled trough the categorical distributions $\lambda_{i_1,\dots,i_L}(j) = P(Q_u=j \mid z_{ch_1(u)} = i_1, \dots, z_{ch_L(u)}=i_L)$ and $\kappa^l_j(i) = P^l(z_u=i \mid Q_u = j)$. The symbols uses to represent these distributions are intentionally the same as those of the Tucker decomposition parameters since there is a strong connection between them. In fact, if we derive the state-transition distribution $P(Q_u \mid Q_{ch_1(u)}, \dots, Q_{ch_L(u)})$ according to the Markov blanket in Figure \ref{fig:markov_blanket_u}, we obtain:
    \begin{equation}
    \begin{split}
    &P(Q_u = j \mid Q_{ch_1(u)} = j_1, \dots, Q_{ch_L(u)}=j_L) =\\
    &\sum_{i_1=1}^{k_1} \cdots \sum_{i_L=1}^{k_L} P(Q_u = j \mid z_{ch_1(u)} = i_1, \dots, z_{ch_L(u)}=i_L)\\
    &\quad\times \prod_{l=1}^L P( z_{ch_l(u)} = i_l \mid Q_{ch_l(u)}=j_l)=\\
    &\sum_{i_1=1}^{k_1} \cdots \sum_{i_L=1}^{k_L} \lambda_{i_1,\dots,i_L}(j) \prod_{l=1}^L \kappa^{l}_{j_l}(i_l).
    \end{split}
    \label{eq:state_trans_decomp}
    \end{equation}
    
    The decomposition obtained in \eqref{eq:state_trans_decomp} is similar to the one obtained in \eqref{eq:Tucker_decomp}: the only difference is that in the latter one both the core tensor $\boldsymbol{\lambda}$ and the mode matrices $\boldsymbol{\kappa}^{l}$ are probability distributions. The value $k_l$ represents the size of the core tensor $\boldsymbol{\lambda}$ along the $l$-th dimension. This Bayesian counterpart of the Tucker decomposition has been already introduced to approximate the conditional distribution in a classification problem \cite{Yang2016} and the state-transition distribution of High-Order Hidden Markov Model \cite{Sarkar2018}.
    
    To deeply understand the probabilistic interpretation of our approximation, it is worth to observe that the states of $z_{ch_l}$ represent clusters of $Q_{ch_l}$ states. If we denote with $\mathcal{C}^l_{i}$ the cluster associated to the $i$-th value of $z_{ch_l}$, the value $\kappa^l_{j}(i)$ indicates the probability that the state $Q_{ch_l}=j$ belongs to the $i$-th cluster. All the states in the same cluster are indistinguishable from the state-transition point of view since the parent state depends only on the cluster (i.e. the state of $z_{ch_l}$). Hence, each cluster contains all the $Q_{ch_l}$ states that bring the same information to the parent state transition. This interpretation makes clear why the Tucker approximation is able to compress the space required to store the state-transition distribution preserving a probabilistic interpretation.
    
    Finally, using the BHTMM complete likelihood equation in \eqref{eq:BTHMM_like} and the approximation introduced in \eqref{eq:state_trans_decomp}, we can derive the complete likelihood equation for the TF-BHTMM:
    \begin{equation}
    \begin{split}
    &\mathcal{L}(\mathbf{x}, \mathbf{Q}, \mathbf{z} \mid \theta)  = P(\mathbf{x}, \mathbf{Q}, \mathbf{z} \mid \theta) =
    \prod_{u \in \mathcal{LF}} \pi^l(j) \, b_j(x_u)\\
    &\quad \times \prod_{u \in \mathcal{U} \setminus \mathcal{LF}}\lambda_{i_1,\dots,i_L}(j) \prod_{l=1}^L \kappa^{l}_{j_l}(i_l) \, b_j(x_u).    
    \end{split}
    \label{eq:HOSVD-BTHMM_like}
    \end{equation}
    We use the symbol $\theta = \{ \boldsymbol{\pi},\mathbf{b},\boldsymbol{\lambda},\boldsymbol{\kappa^1},\dots,\boldsymbol{\kappa^L} \}$ to denote the set of parameters which define a TF-BHTMM.
    
    \subsection{Decomposition Size}
    The values  $\mathbf{k} = \{k_1, \dots, k_L\}$ (i.e. the size of the core tensor $\boldsymbol{\lambda}$) play a fundamental role in the compression of the state-transition distribution. In fact, as we discussed in Section \ref{sec:tucker_decomp}, the space required to store the Tucker factorisation depends on the size of the core tensor. In our case, it is $C \times \prod_{l=1}^L k_l$. Moreover, these quantities have a nice probabilistic interpretation \cite{Sarkar2018}: the value $k_l$ measures how strong is the dependence between the hidden state $Q_{ch_l}$ and its hidden parent state $Q_u$. In fact, a value of $k_l=C$, means that it is important to know the exact child state $Q_{ch_l}$ in order to determine the parent state. On the contrary, if $k_l =1$, the child state does not affect the parent one: no matter the true value of $Q_{ch_l}$, it will collapse in the unique state available for $z_{ch_l}$.
    
    For ease of exposition, in Section \ref{sec:model_intro}, we have assumed the true size of the core tensor to be known. In practical application, these quantities are unknown. To overcome this limitation, we treat them as discrete hidden random variables, hence the Bayesian nature of our model. To this end, we specify a priori distribution on them and we learn the most suitable values during the training procedure.
    
    The prior distributions on such variables are very important since they force the model to focus either on compression (i.e. the $\mathbf{k}$ values are almost $1$) or data representation (i.e. the $\mathbf{k}$ values are almost $C$). Moreover, we can easily insert prior knowledge: for example, we can build a prior distribution which is position dependent to indicate there are children positions that are more informative than the others.
    
    In our model, we decide to use the following position-independent priori:
    \begin{align}
    p(k_l) &= e^{-\varphi k_l}, \label{eq:exp_priori}\\
    \sum_{l=1}^L \mathbb{I}[k_l \neq 1] &\in [L_{\text{min}}, L_{\text{max}}] \label{eq:size_constraints}.
    \end{align}
    The value $\varphi$ is an hyper-parameter which regulates the distribution decay. The hyper-parameters $L_{\text{min}}$ and $L_{\text{max}}$ denote the minimum and the maximum number of important positions, respectively. The utility of $L_{\text{min}}$ and $L_{\text{max}}$ is twofold: (i) we can avoid that the parent hidden state is completely independent from the child hidden states (by setting $L_{\text{min}} \geq 1$), and (ii) we can control the worst case space-complexity (which is $O(C^{L_{\text{max}}+1})$).
    
    By varying the value of $k_l$, we change the dimension of the core tensor $\boldsymbol{\lambda}$. Hence, we need to specify a prior distribution on it in order to generate new entries on the fly. Since $\boldsymbol{\lambda}$ is a categorical distribution, we use its conjugate prior, i.e. the Dirichlet distribution
    \begin{align}
    \lambda_{i_1,\dots,i_L} &\sim \text{Dirichlet}(\alpha \boldsymbol{\lambda_0})\\
    \boldsymbol{\lambda_0} &\sim \text{Dirichlet}(\alpha_0/C,\dots,\alpha_0/C).
    \end{align}
    The value $\alpha$ and $\alpha_0$ are hyper-parameters which regulate the shape of the $\boldsymbol{\lambda}$ distribution.
    
    \subsection{Parameter Learning}
    Learning of TF-BHTMM parameters is straightforward if the vector dimension $\mathbf{k}$ is known. As we discussed in the previous section, in practical application these quantities are unknown. Therefore, we develop a learning procedure which is able to learn the approximation size from data. 
    
    Unfortunately, the estimation of the vector $\mathbf{k}$ worsens the learning problem. To this end, we introduce an approximation on the mode matrices $\boldsymbol{\kappa^1},\dots,\boldsymbol{\kappa^L}$. The approximation consists in adding the constraint $\tilde{\kappa}^l_{j}({i})=\{0,1\}$ to each entry of the mode matrices \cite{Sarkar2018}. This new constraint makes the clustering discussed in Section \ref{sec:model_intro} deterministic. In fact, we can associate each state to a single cluster and define each cluster as $\mathcal{C}^l_i = \{j \mid \tilde{\kappa}^l_{j}({i})=1\}$. The approximated mode matrices $\tilde{\mathbf{\kappa}}^l$ take the name of \textit{hard clustering}, which is opposed to the \textit{soft clustering} defined by the mode matrices without approximation.
    
    The approximation above allows us to develop a Gibbs sampling algorithm for fitting the unknown quantities from data. The procedure comprises the following steps:
    \begin{enumerate}
        \item update the hidden state $\mathbf{Q}, \mathbf{z}$;
        \item update the approximation size $\mathbf{k}'$ and the mode matrices $\boldsymbol{\kappa^1},\dots,\boldsymbol{\kappa^L}$;
        \item update the model parameters $\boldsymbol{\pi},\boldsymbol{\lambda},\boldsymbol{\lambda_0},\mathbf{b}$.
    \end{enumerate}
    For the sake of simplicity, we discuss the learning algorithm restricting to  training set with a single tree. The extension to a generic i.i.d. dataset is straightforward.
    
    In the first step we perform a Simulated Annealing \cite{Kirkpatrick1983} update of latent variables. Given the current values of $(\mathbf{Q},\mathbf{z},\mathbf{k},\theta)$, we compute the new values of $\mathbf{Q}',\mathbf{z}'$ through an ancestor sampling procedure \cite{Barber2016} starting from leaf nodes. In particular, we sample the new $\mathbf{Q}'$ fixing the old values $\mathbf{z}$. Then, we use $\mathbf{Q}'$ to sample the new values $\mathbf{z}'$. More details are given in Algorithm \ref{alg:latent_sampling}. The new values $\mathbf{Q}',\mathbf{z}'$ are accepted with probability \cite{Sarkar2018}
    \begin{equation}
    \min\left\{\left[\frac{\prod_u \lambda_{z'_l}(Q'_u)}{\prod_u \lambda_{z_l}(Q_u)}\frac{\prod_u \lambda_{z'_l}(Q_u)}{\prod_u \lambda_{z_l}(Q'_u)}\right]^{1/\mathcal{T}(m)},1\right\},
    \end{equation}
    where $\mathcal{T}(0)$ and $\mathcal{T}(m) = \max\{\mathcal{T}_0^{1-m/m_0}, 1\}$ denotes the initial and the current annealing temperature, $m$ is the current iteration number and $m_0$ is the iteration at which the temperature reduces to one. 
    
    \begin{algorithm}[tb]    
        \caption{Sample latent states $\mathbf{Q}', \mathbf{z}'$}\label{alg:latent_sampling}
        \begin{algorithmic}[1]
            \ForAll{$u \in \mathcal{LF}(t)$} \Comment{Iterate over leaf nodes}
            \State $Q'_u \sim \pi^l$, where $l=pos(u)$
            \EndFor
            \ForAll{$u \notin \mathcal{LF}(t)$}\Comment{Iterate over internal nodes}
            \For{$l=1 \textbf{ to } L$}
            \State $j = Q'_{ch_l(u)}$\Comment{Bottom if the child does not exist}
            \State $z'_{ch_l(u)} \sim \tilde{\kappa}^l_{j}$
            \EndFor
            \State $Q'_u \sim \lambda_{z_{ch_1},\dots,z_{ch_L}}$
            \EndFor
        \end{algorithmic}
    \end{algorithm}
    
    In the second step, we perform a Stochastic Search for Variable Selection (SSVS) \cite{George1997} to update the compression size. The approximation size $\mathbf{k}$ is updated choosing a random position $l$. Then, we decide to do an increase (or decrease) move on the position $l$ by a coin toss. The increasing move consists in adding a new cluster for the $l$ position. Hence, we randomly select a cluster $\mathcal{C}^l_i$ and we randomly split it into two clusters. The splitting process consists to randomly select a set of element in the cluster $\mathcal{C}^l_i$ and to move them in the new cluster $\mathcal{C}^l_{i'}$. In practice, it is implemented modifying the hard clustering $\tilde{\mathbf{\kappa}}^l$: for every state $j$ which is moving from cluster $i$ to cluster $i'$, we set $\tilde{\kappa}^l_j(i) = 0$ and $\tilde{\kappa}^l_j(i') = 1$. On the contrary, the decrease move merges together two randomly selected clusters $\mathcal{C}^l_i$ and $\mathcal{C}^l_{i'}$. Again, the merging operation is implemented modifying the hard clustering $\tilde{\mathbf{\kappa}}^l$: for every state $j$ in the cluster $i'$, we set $\tilde{\kappa}^l_j(i') = 0$ and $\tilde{\kappa}^l_j(i) = 1$.
    
    Finally, we should guarantee that the constraint \eqref{eq:size_constraints} is satisfied: Algorithm \ref{alg:size_sampling} provides more details on the process. The new values $\mathbf{k}'$  are accepted with probability \cite{Sarkar2018}
    \begin{equation}
    \min\left\{\left[\frac{\mathcal{L}(\mathbf{k}')}{\mathcal{L}(\mathbf{k})}\frac{\prod_lp(k'_l)}{\prod_lp(k_l)}\right]^{1/\mathcal{T}(m)},1\right\},
    \end{equation}
    where $\mathcal{T}(m)$ is the current annealing temperature which is computed as in the previous step. The priori $p(k_l)$ is defined in \eqref{eq:exp_priori}, while the marginal likelihood is given by \cite{Sarkar2018}:
    \begin{equation}
    \mathcal{L}(\mathbf{k}) = \prod_{\mathbf{i_u}} \frac{\mathbf{B}(\alpha\lambda_0(1) + n_\mathbf{i_u}(1),\dots,\alpha\lambda_0(C)+n_\mathbf{i_u}(C))}{\mathbf{B}(\alpha\lambda_0(1),\dots,\alpha\lambda_0(C))},
    \end{equation}
    where $\mathbf{i_u} = (i_1,\dots,i_L)$.    The function $\mathbf{B}(\cdot)$ represents the multivariate Beta function and the value $n_\mathbf{i_u}(j)$ counts how many times the joint configuration $(Q_u =j, Q_{ch_1} = i_1,\dots,Q_{ch_L}=i_l)$ appears.
    
    \begin{algorithm}[tb]
        \caption{Sample size vector $\mathbf{k}$}\label{alg:size_sampling}
        \begin{algorithmic}[1]
            \State $l \sim Uniform(L)$\Comment{Choose the position}            
            \State $v \sim Uniform(2)$\Comment{Random move}
            \If{$k_l=1$} \Comment{Must do increase move}
            \State $v=1$
            \EndIf
            \If{$k_l=C$} \Comment{Must do decrease move}
            \State $v=2$
            \EndIf
            
            \If{$v=1$}\Comment{Do increase move}
            \State $k_l = k_l+1$ 
            \State Randomly split a random cluster, modifying $\boldsymbol{\tilde{\kappa}^l}$
            \Else\Comment{Do decrease move}
            \State $k_l = k_l-1$ 
            \State Merge two random clusters, modifying $\boldsymbol{\tilde{\kappa}^l}$
            \EndIf
            \If{$\sum_{p=1}^L \mathbb{I}[k_p \neq 1] > L_{\text{max}} $} \Comment{Check constraints}
            \State Randomly select a position $l'$ s.t. $k_{l'}>1$
            \State Do the decrease move on position $l'$
            \If{$\sum_{p=1}^L \mathbb{I}[k_p \neq 1] > L_{\text{max}} $}
            \State Remove the increase move on $l$
            \EndIf
            \EndIf
            
            \If{$\sum_{p=1}^L \mathbb{I}[k_p \neq 1] < L_{\text{min}} $} \Comment{Check constraints}
            \State Randomly select a position $l'$ s.t. $k_{l'}=1$
            \State Do the increase move on position $l'$
            \EndIf
        \end{algorithmic}
    \end{algorithm}
    
    In the last step, we update the model parameters sampling from their multinomial conditional. We assume that both the priori distribution $\boldsymbol{\pi}$ and the emission distribution $\mathbf{b}$ have a Dirichlet priori with parameters $\gamma$ and $\beta$, respectively, yielding     
    \begin{align}
    \pi^l &\sim \text{Dirichlet}(\gamma + n^l(1),\dots,\gamma + n^l(C))\\
    b_j &\sim \text{Diricchlet}(\beta + n_j(d_1),\dots,\beta + n_j(d_M))\\
    \lambda_{\mathbf{i_u}} &\sim \text{Dirichlet}(\alpha\lambda_0(1) + n_\mathbf{i_u}(1),\dots,\alpha\lambda_0(C)+n_\mathbf{i_u}(C)),
    \end{align}
    where $n^l(c) = \sum_{u\in\mathcal{LF}} \mathbb{I}[Q_u=c \, \wedge \, pos(u)=l]$ and $ n_j(d) = \sum_u \mathbb{I}[Q_u=j \wedge x_u=d]$. The operator $\mathbb{I}$ is the indicator function whose values is $1$ if and only if the condition in its square brackets is true. 
    
    The sampling of the base distribution $\boldsymbol{\lambda_0}$ requires a more complex procedure which is outlined in Algorithm \ref{alg:lamb_zero_sampling}.

    \begin{algorithm}[tb]
        \caption{Sample vector $\boldsymbol{\lambda_0}$}\label{alg:lamb_zero_sampling}
        \begin{algorithmic}[1]
            \ForAll{$\mathbf{i} =(i_1,\dots,i_L)$}
            \For{$p=1 \textbf{ to } n_{\mathbf{i}}(c)$}
            \State $x_p \sim Bernoulli\left(\frac{\alpha\lambda_0(c)}{p-1+\alpha\lambda_0(c)}\right)$
            \EndFor
            \State $m_{\mathbf{i}}(c) = \sum_p x_p$
            \EndFor
            \State $m_0(c) = \sum_{\mathbf{i}} m_{\mathbf{i}}(c)$
            \State $\boldsymbol{\lambda_0} \sim \text{Dir}(\alpha_0/C,\dots + m_0(1),\alpha_0/C+m_0(C))$
        \end{algorithmic}
    \end{algorithm}
    
    \section{Experimental Analysis}\label{sec:experiment}
    In this section, we provide an experimental evaluation of the proposed model, with the aim of characterising the properties of the approximation introduced against the SP one. In particular, we evaluate the model on two different tasks: (1) a classification task on XML tree-data and (2) a labelling task on a synthetic dataset. In both tasks, we use two measures to evaluate the model performance: the \textit{accuracy}, which assess the correctness of model's answers, and the \textit{entropy}, which measures the amount of uncertainty in them. For the accuracy measures, higher values are better; for the entropy, lower values are better.
    
    All the reported results are averaged over five executions, to account for randomisation effects due to initialisation. The MATLAB code used can be found on a public GitLab repository\footnote{https://gitlab.itc.unipi.it/d.castellana/TF\_bhtmm}.
    
    \subsection{Classification Task}
    The classification task consists in predicting the class a tree-structured sample belongs to. We test both the SP-BHTMM and TF-BHTMM on two datasets taken from the INEX 2005 (INEX05) and INEX 2006 (INEX06) competition \cite{Denoyer2008}. The INEX05 dataset is based on the (m-db-s-0) corpus, comprising $9631$ XML-formatted documents represented as trees with maximum output degree $L=32$ and assigned to $11$ different cluster. Node labels represent $366$ different XML tags. The dataset is split in training set ($4820$ trees) and test set ($4811$ trees) \cite{Denoyer2008}. The INEX06 is composed of $12107$ XML- formatted documents representing scientific articles, each from one of $18$ different IEEE journals which represent the different cluster. Again, the node labels represent different $65$ XML tags and maximum output degree $L=66$. The dataset  is split in training set ($6,053$ trees) and test set ($6,054$ trees) \cite{Denoyer2008}.
    
    On both datasets, we train a single model for each class. Each model is trained on the elements in the training set associated with the same model class; at test time, we compute the likelihood according to each model and we assign the class of the model which scores the highest sample likelihood. For all the models, we set a flat prior on model parameters. Also we set $\varphi=2$ (as in \cite{Sarkar2018}), $L_{\text{min}} = 1$, and $L_{\text{max}} = 5$ for the TF-BHTMM. The values assigned to $L_{\text{min}}$ and $L_{\text{max}}$ allow to bound the space complexity of the model to $O(C^6)$.
    
    To evaluate the impact of the hidden state size, we vary the number of hidden states $C \in \{2,4,6,8,10\}$. Both training algorithms are executed for $100$ iterations.
    
    In Table \ref{tab:INEX05_results} we report the accuracy and the entropy obtained on the test set. The results show that the TF-BHTMM always outperforms the SP-BHTMM, both in accuracy and in entropy: the difference is higher when the number of hidden states is small. This results is quite surprising if we consider that the TF-BHTMM considers at most $5$ child nodes among $32$ (we set $L_{\text{max}} = 5$); on the contrary, the SP-BHTMM always consider the contribution of all child nodes. Hence, we deduce that not all the child nodes contain useful information for the tree-classification and therefore can be ignored.    
    
    \begin{table}[tb]
        \centering    
        \resizebox{\columnwidth}{!}{
            \begin{tabular}{c|cc|cc}
                \hline
                & \multicolumn{4}{ |c }{INEX05} \\
                \hline
                \multirow{2}{*}{Hidden} & \multicolumn{2}{ |c| }{Accuracy ($\%$)} & \multicolumn{2}{ |c }{Entropy ($\%$)} \\
                & SP-BU & TU-BU & TF-BU & TU-BU\\
                \hline
                $C=2$ & $87.55\;(4.13)$ & $\mathbf{91.41}\;(3.62)$ & $34.19\;(5.80)$ & $ \mathbf{31.99}\;(10.37)$\\
                $C=4$ & $90.68\;(5.70)$ & $\mathbf{93.65}\;(1.64)$ & $28.92\;(5.16)$ & $\mathbf{25.00}\;(4.13)$\\
                $C=6$ & $93.81\;(1.15)$ & $\mathbf{95.10}\;(0.27)$ & $24.30\;(2.90)$ & $\mathbf{20.91}\;(0.93)$\\
                $C=8$ & $93.15\;(1.69)$ & $\mathbf{94.28}\;(2.27)$ & $23.59\;(1.29)$ & $\mathbf{23.12}\;(6.35)$\\
                $C=10$ & $93.30\;(3.09)$ & $\mathbf{95.21}\;(0.17)$ & $21.77\;(1.64)$ & $\mathbf{20.60}\;(0.46)$\\
                \hline
            \end{tabular}
        }
        \caption{Average accuracy and entropy over 5 runs (std in brackets) on INEX05 dataset. In bold the best result for each model}
        \label{tab:INEX05_results}
    \end{table}

    The results obtained on the INEX06 dataset (see Table \ref{tab:INEX06_results}) are similar to the ones obtained on the INEX05 dataset: the TF-BHTMM always reach a higher accuracy than the SP-BHTMM. On the contrary, the entropy values are high for both model due to the intrinsic difficulty of the INEX06 dataset \cite{Bacciu2012}.
    \begin{table}[tb]
        \centering    
        \footnotesize
        \resizebox{\columnwidth}{!}{
            \begin{tabular}{c|cc|cc}
                \hline
                & \multicolumn{4}{ |c }{INEX06} \\
                \hline
                \multirow{2}{*}{Hidden} & \multicolumn{2}{ |c| }{Accuracy ($\%$)} & \multicolumn{2}{ |c }{Entropy ($\%$)} \\
                & SP-BU & TU-BU & SP-BU & TF-BU\\
                \hline
                $C=2$ & $21.44\;(4.54)$ & $\mathbf{27.94}\;(2.62)$ & $287.40\;(7.24)$ & $275.28\;(2.07)$\\
                $C=4$ & $24.84\;(3.15)$ & $\mathbf{29.83}\;(3.06)$ & $281.23\;(3.46)$ & $279.82\;(7.11)$\\
                $C=6$ & $25.57\;(2.12)$ & $\mathbf{30.65}\;(2.36)$ & $277.99\;(2.81)$ & $283.47\;(5.34)$\\
                $C=8$ & $26.43\;(2.47)$ & $\mathbf{26.48}\;(2.84)$ & $278.17\;(1.12)$ & $292.36\;(5.56)$\\
                $C=10$ & $22.89\;(3.33)$ & $\mathbf{26.94}\;(2.77)$ & $289.40\;(6.28)$ & $291.03\;(7.12)$\\
                \hline
            \end{tabular}
        }
        \caption{Average accuracy and entropy over 5 runs (std in brackets) on INEX06 dataset. In bold the best result for each model}
        \label{tab:INEX06_results}
    \end{table}
    
    \subsection{Labelling Task}
    The labelling task consists in predicting the visible label associated with the nodes of a given tree structure. We test both the SP-BHTMM and the TF-BHTMM on a controlled dataset which contains ternary trees (i.e. L=3), comprising left-asymmetric, symmetric and right-asymmetric tree. A tree is defined as left-asymmetric (right-asymmetric) if the number of nodes in the leftmost (rightmost) position is greater than the number of nodes in the opposite position. In a symmetric tree, the number of nodes is almost equivalent for each position. The label of each node encodes structural information since it represents the number of children of the node: therefore the label goes from $0$ (i.e. no child nodes) to $3$ (i.e. a child node in each position). We generate $780$ trees ($260$ for each type) and split them in training set ($600$ trees, $200$ for each type) and test set ($180$ trees, $60$ for each type).
    
    We train a SP-BHTMM and a TF-BHTMM with $C=10$ hidden states. In this experiments, we do not test multiple hyper-parameter configurations: for both models, we use flat priors to generate the initial probability distribution. Also we set $\varphi=2$ (as in \cite{Sarkar2018}), $L_{\text{min}} = 1$, and $L_{\text{max}} = 3$ for the TF-BHTMM. The values assigned to $L_{\text{min}}$ and $L_{\text{max}}$ allow to consider all child position. Both training algorithms are executed for $100$ iteration. At test time, we remove the visible labels from the test data and we generate them using the probability distribution learned by both models. In Table \ref{tab:labelling_results}, we report accuracy and entropy obtained by both models on each label type.
    
    \begin{table}[tb]
        \centering
        \resizebox{\columnwidth}{!}{
            \begin{tabular}{c|cc|cc}
                \hline
                & \multicolumn{4}{|c}{Synthetic dataset}\\
                \hline
                \multirow{2}{*}{Label} & \multicolumn{2}{|c}{Accuracy ($\%$)} & \multicolumn{2}{|c}{Entropy ($\%$)}\\
                & SP-BU & TU-BU & SP-BU & TF-BU\\
                \hline    
                $0$ & $55.47\;(12.90)$ & $\mathbf{99.67}\;(0.11)$ & $35.49\;(8.25)$ & $\mathbf{1.34}\;(2.03)$\\
                $1$ & $60.15\;(2.22)$ & $\mathbf{79.59}\;(22.35)$ & $151.51\;(13.14)$ & $\mathbf{65.61}\;(61.25)$\\
                $2$ & $45.84\;(4.43)$ & $\mathbf{64.71}\;(27.17)$ & $180.88\;(5.40)$ & $\mathbf{93.69}\;(51.19)$\\
                $3$ & $17.04\;(2.05)$ & $\mathbf{29.68}\;(38.22)$ & $184.08\;(4.93)$ & $\mathbf{97.71}\;(44.38)$\\
                \hline
                All & $53.08\;(4.83)$ & $\mathbf{80.68}\;(15.77)$ & $140.38\;(10.39)$ & $\mathbf{51.42}\;(34.91)$\\
                \hline
            \end{tabular}
        }
        \caption{Average label accuracy over 5 runs (std in brackets) on the synthetic dataset. In bold the best result for each model}
        \label{tab:labelling_results}
    \end{table}
    
\begin{figure}[tb]
        \centering
        \begin{subfigure}[t]{0.48\columnwidth}
            \centering      
            \includegraphics[width=\textwidth]{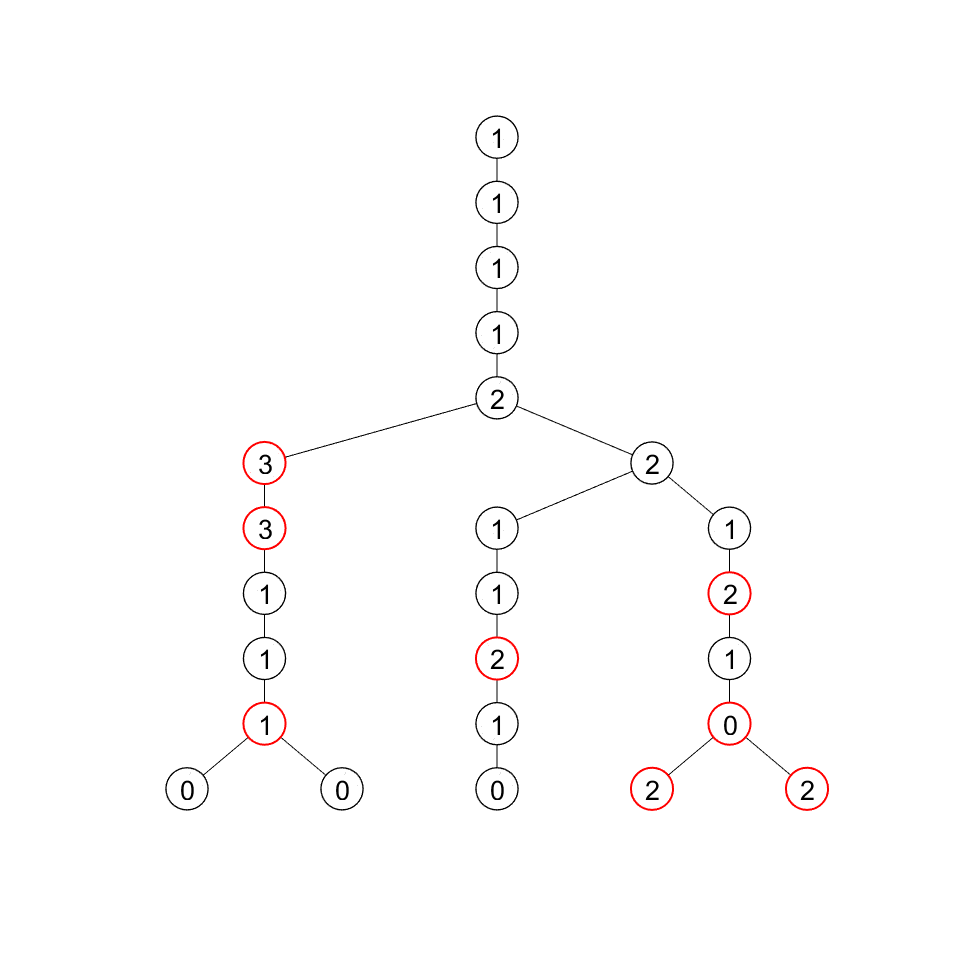}
            \caption{Labels generated by the SP-BHTMM.}
            \label{fig:SP_example}
        \end{subfigure}
        \begin{subfigure}[t]{0.48\columnwidth}
            \centering      
            \includegraphics[width=\textwidth]{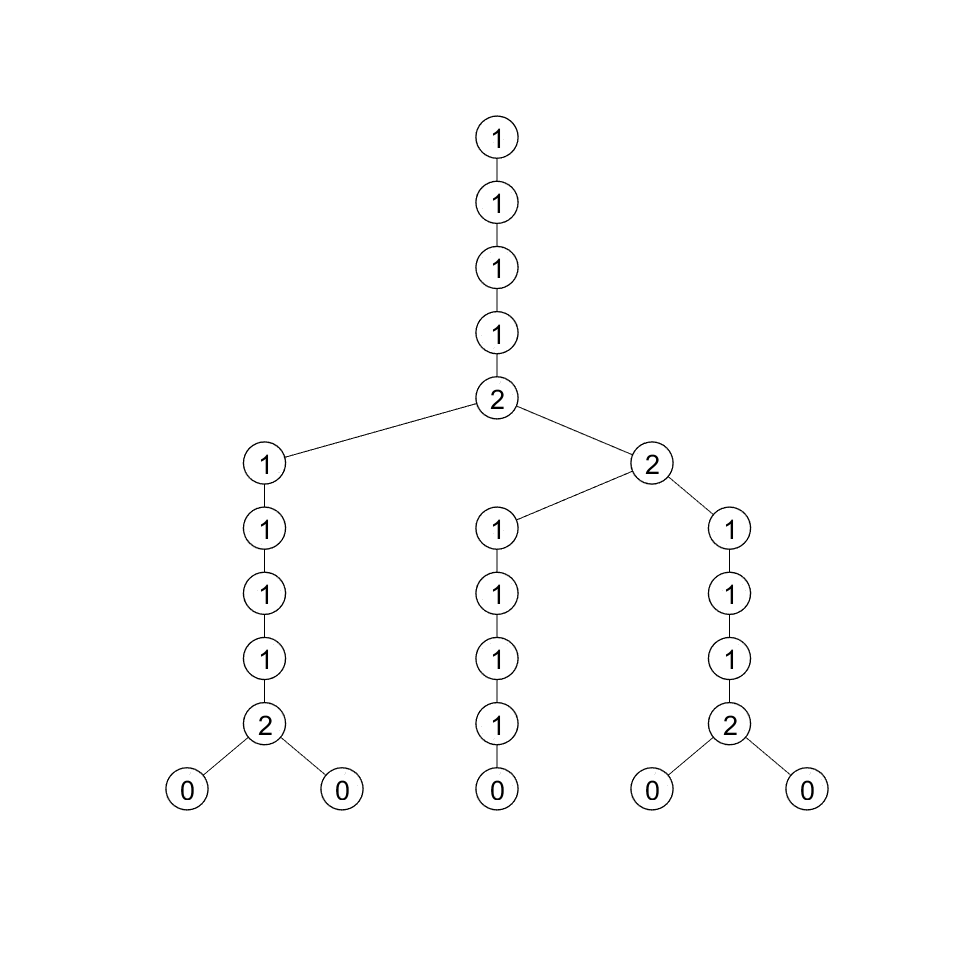}
            \caption{Labels generated by the TF-BHTMM.}
            \label{fig:SVD_example}
        \end{subfigure}
        \caption{An example of tree label generation on the synthetic dataset by SP-BHTMM and TF-BHTMM best execution. Red nodes have wrong label.}
        \label{fig:labelling_example}
    \end{figure}

    The overall accuracy reached by SP-BHTMM and TF-BHTMM is  $53.08\%$ and $80.68\%$ respectively, showing the effectiveness of the approximation introduced. Even if the prediction of the label $0$ should be the easiest one (it appears only on leaf nodes), the accuracy obtained by the SP-BHTMM is only around $55\%$ while the accuracy of TF-BHTMM is around $99\%$. The high entropy value suggests that the SP-BTHMM attaches the $0$ label also to internal nodes (see Figure \ref{fig:SP_example}).
    
\begin{figure}[tb]
        \centering
        \begin{subfigure}[t]{0.48\columnwidth}
            \centering      
            \includegraphics[width=\textwidth]{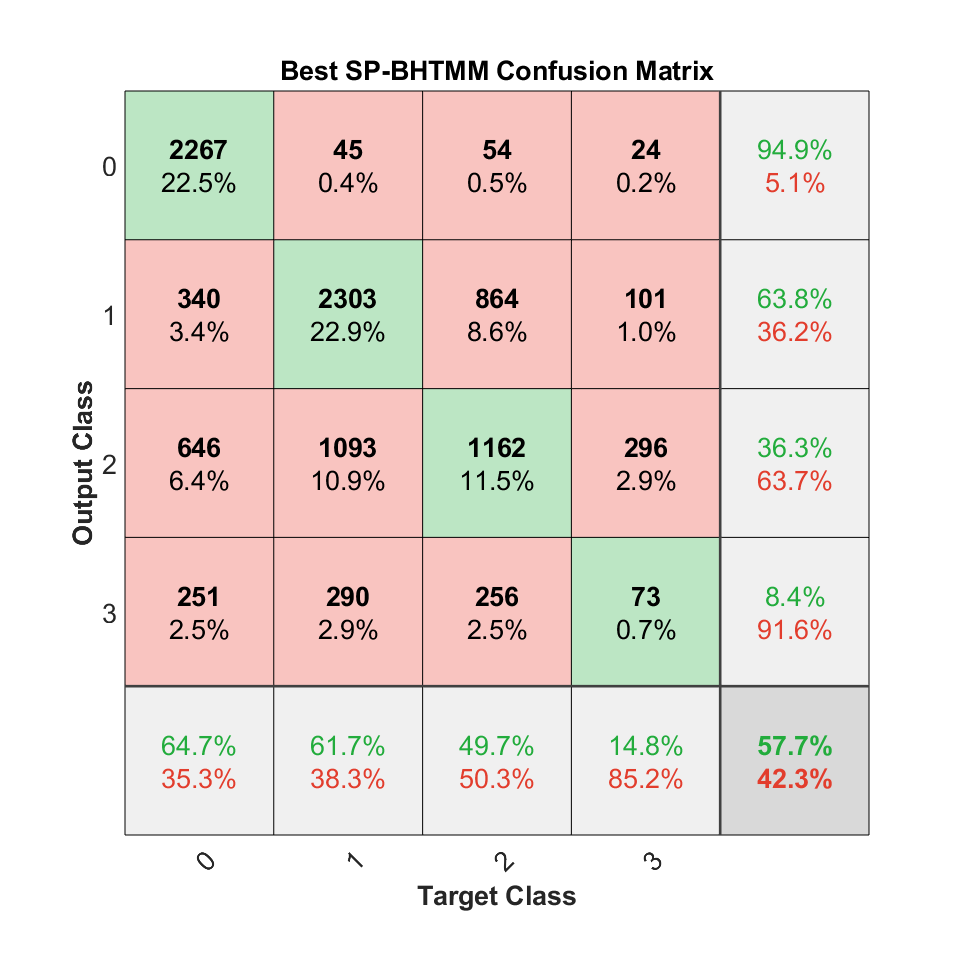}
            \caption{SP-BHTMM confusion matrix.}
            \label{fig:SP_confmat}
        \end{subfigure}
        \begin{subfigure}[t]{0.48\columnwidth}
            \centering      
            \includegraphics[width=\textwidth]{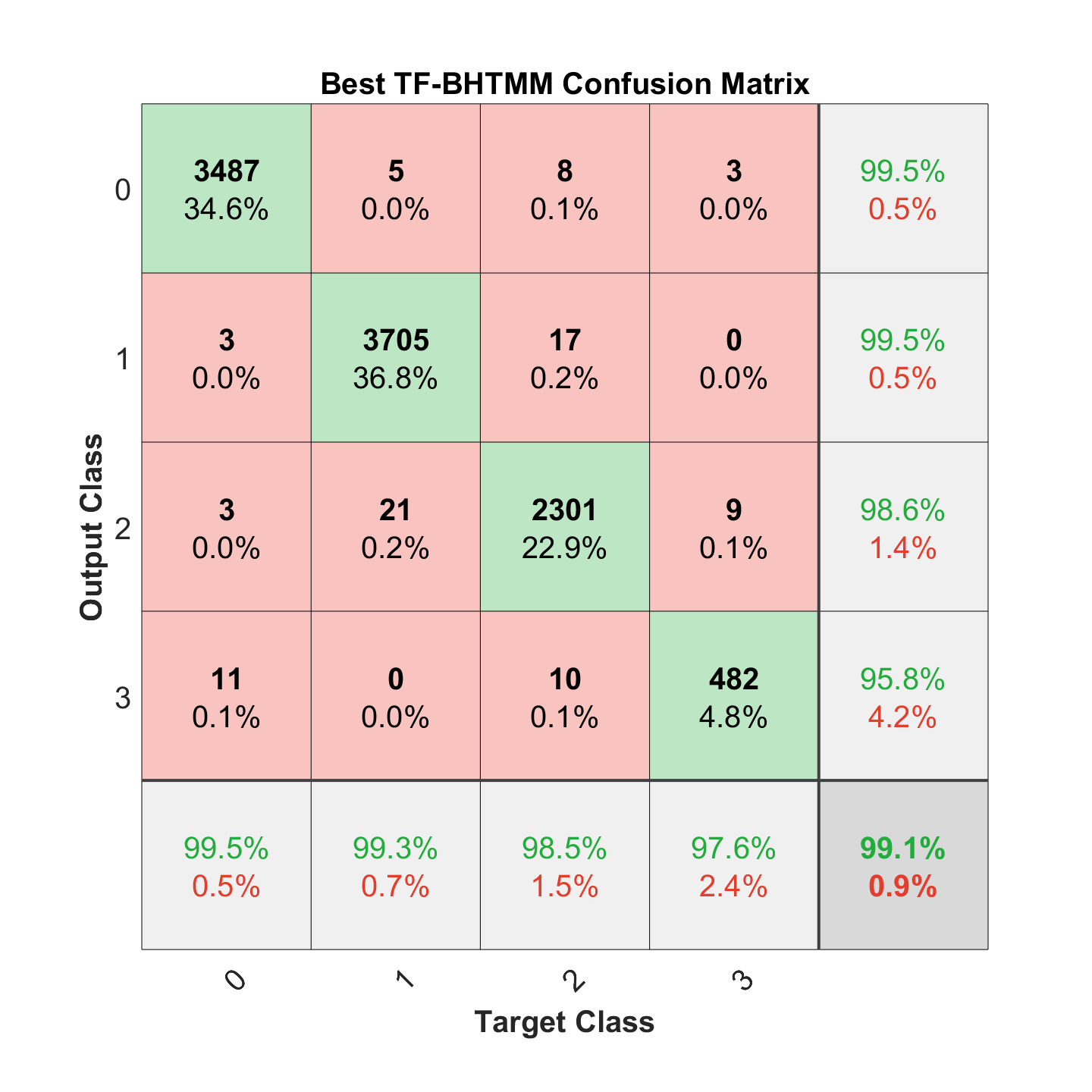}
            \caption{TF-BHTMM confusion matrix.}
            \label{fig:SVD_confamt}
        \end{subfigure}
        \caption{Confusion matrices obtained by SP-BHTMM and TF-BHTMM best execution on synthetic dataset.}
        \label{fig:labelling_confmat}
    \end{figure}

    In Figure \ref{fig:labelling_confmat} we report the confusion matrix obtained by the best SP-BHTMM and the best TF-BHTMM. The best TF-BHTMM reaches an accuracy of $99.1\%$, while the best SP-BHTMM reaches only an accuracy of $57.5\%$. This huge difference is due to the SP approximation which mixes together the contributions from child nodes. On the contrary, the TF-BHTMM is able to distinguish the contribution of each child due to the core tensor $\boldsymbol{\lambda}$ which models all the possible joint configurations of the $z$ variables on child nodes.
    
    Moreover, there is consistent difference between the accuracy obtained by the best TF-BHTMM model and the results obtained averaging TF-BHTMM over $5$ runs. This is confirmed by the high standard deviation reported in Table \ref{tab:labelling_results}. We believe that the high variance in the results is due to strong dependence between the first two step of the learning algorithm.
    
    \section{Conclusion}\label{sec:conclusion}
    In this paper, we have introduced a new generative model which defines a probability distribution over tree-structured data. The novelty of our model is the approximation introduced to tackle the well-known computational problems which arise in BU-HTMMs due to the exponential nature of the bottom-up state transition distribution. We have also introduced a new training algorithm which is able to learn the approximation size directly from data, simplifying the model selection step. Finally, we have shown the potential of our model by achieving better results than SP-BHTMM in two different tasks. In particular, the results obtained on the labelling task show clearly the advantage given by the new approximation.
    
    The emphasis of this paper is on the approximation which exhibits nice property and promising results to further develop the model. One of the first improvement concerns the learning algorithm, which exhibits a high variance on the labelling task. Further study should be conducted in order to reduce the dependence between the first two steps of the Gibbs sampler.
    
    Also, the model can be naturally extended in a Bayesian non-parametric fashion, allowing to learn the size of its latent space directly from data. Thanks to the approximation, the size of the latent space does not affect the complexity of the model which is regulated by the approximation size. In this sense, the non-parametric model will be able to separate the label generation dynamics, which depends on the size of the latent space, from the state-transition dynamic, which depends on the approximation size.
    
\section*{Acknowledgment}
The work is supported by the Italian Ministry of Education, University, and Research (MIUR) under project SIR 2014 LIST-IT (grant n. RBSI14STDE).
    
    \FloatBarrier
    
    \bibliography{biblio}
    
\end{document}